\documentclass[10pt,twoside]{elsart}
\usepackage{amsmath}

\usepackage{graphicx}
\usepackage[bf,centerlast]{subfigure}

\numberwithin{equation}{section}

\newcommand{\F}{\mathbf{F}}
\newcommand{\Hb}{\mathbf{H}}

\newcommand{\x}{\mathbf{x}}
\newcommand{\y}{\mathbf{y}}

\newcommand{\K}{\mathbf{K}}

\newcommand{\n}{\mathbf{n}}

\newcommand{\w}{\mathbf{w}}

\newcommand{\m}{\mathbf{m}}
\newcommand{\vb}{\mathbf{v}}
\newcommand{\M}{\mathbf{M}}

\newcommand{\N}{\mathcal{N}}

\begin{document}
\begin{frontmatter}

\author{B. P\'oczos}
\author{A. L{\H o}rincz\corauthref{cor1}}
\address{Department of Information Systems, E\"otv\"os Lor\'and
University, P\'azm\'any P\'eter s\'et\'any 1/C, Budapest
Hungary,H-1117} \corauth[cor1]{corresponding author Address:
Department of Information Systems, E\"otv\"os Lor\'and University,
P\'azm\'any P\'eter s\'et\'any 1/C, Budapest Hungary, H-1117
P:36-1-2090555 ext 8483} \ead{lorincz@inf.elte.hu}

\title{Kalman-filtering using local interactions}

\date{\today}%
\maketitle

\begin{abstract}
There is a growing interest in using Kalman-filter models for
brain modelling. In turn, it is of considerable importance to
represent Kalman-filter in connectionist forms with local Hebbian
learning rules. To our best knowledge, Kalman-filter has not been
given such local representation. It seems that the main obstacle
is the dynamic adaptation of the Kalman-gain. Here, a
connectionist representation is presented, which is derived by
means of the recursive prediction error method. We show that this
method gives rise to attractive local learning rules and can adapt
the Kalman-gain.
\end{abstract}

\journal{Journal of Machine Learning Research}
\end{frontmatter}

\section{Introduction}

Linear dynamical systems (LDS) are well studied and widely applied
tools in both state estimation and control. Inference in LDS
becomes simple, unbiased and has minimized covariance if the
Kalman-filter recursion is used. Recently, there is growing
interest in Kalman-filters or Kalman-filter like structures as
models for neurobiological substrates. It has been suggested that
Kalman-filtering (i) may occur at sensory processing
(\cite{rao97dynamic,rao99predictive}), (ii) may be the underlying
computation of the hippocampus (\cite{bousquet99is}), and may be
the underlying principle in control architectures
(\cite{todorov02optimal}). Detailed architectural similarities
between Kalman-filter and the entorhinal-hippocampal loop as well
as between Kalman-filters and the neocortical hierarchy have been
described recently
(\cite{lorincz00parahippocampal,lorincz02mystery}). Interplay
between the dynamics of Kalman-filter-like architectures and
learning of parameters of neuronal networks has promising aspects
for explaining known and puzzling phenomena, such as priming,
repetition suppression and categorization
(\cite{lorincz02relating,keri02categories}).

Kalman-filter is an on-line recursive algorithm. Unfortunately,
Kalman-filtering requires the computation of the Kalman-gain.
Recursions of the Kalman-gain matrix assume that covariance
matrices of measurement noise and observation noise are known. In
general, these parameter sets are not known in advance and may be
subject to temporal changes. Moreover, to determine the
Kalman-gain, the algorithm requires the inversion of matrices,
which is hard to interpret in neurobiological terms. To our best
knowledge, all suggested networks computed the Kalman-gain matrix
directly, e.g., using matrix inversions.

Here, an alternative route, the recursive prediction error (RPE)
method (\cite{Ljung83Theory}) is followed. Using this method, we
were able to construct a special parametrization, which is (i)
on-line and recursive and (ii) makes use of local interactions to
estimate the filtering parameters, including the Kalman-gain. The
next section (Section \ref{s:background}) reviews background
materials, such as the constraints on connectionist systems
(Section \ref{ss:constraints}), the well known Kalman-filer
recursion (Section \ref{ss:KFR}). In Section \ref{ss:RPE} the
recursive prediction error method is applied to estimate the
Kalman-gain. Our particular parametrization and the corresponding
architecture are provided in Section \ref{s:LKF} and Section
\ref{ss:archi}, respectively. Conclusions are drawn in the last
section (Section \ref{s:conc}). Detailed mapping to the neural
substrate is not aimed here: There should be large differences if
the goal is the mapping (i) to the control system of the brain,
(ii) to the hippocampus, (iii) to the hippocampal-entorhinal loop,
or, (iv) to the neocortical layers, etc. All suggestions on
Kalman-filtering need to map \textit{a} connectionist architecture
to \textit{a} part of the brain.

\section{Background} \label{s:background}

\subsection{Constraints on connectionist systems}
\label{ss:constraints}

\begin{figure}[h!]
\centering\includegraphics[width=6cm]{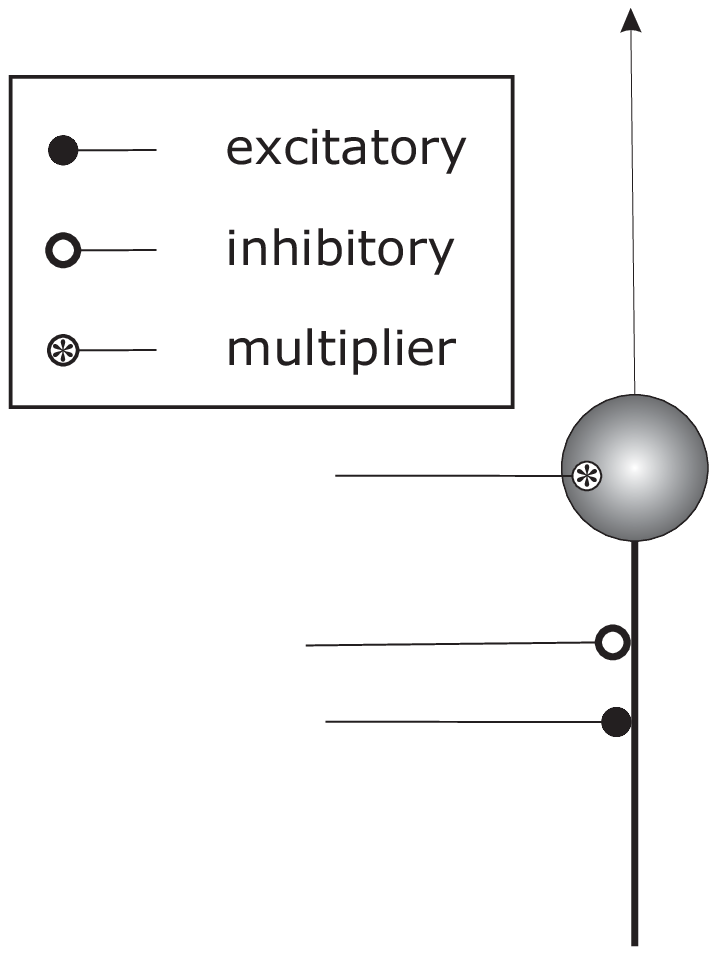}
\caption{\textbf{Local interactions of neurons}
\newline
The node (or neuron) is depicted by the large grey circle.
Connections to and from other neurons are depicted by lines, these
are called directed connections. The direction of a connection is
denoted either by an arrow or by a small circle. The small circle
is the synapse of the connection at around the targeted neuron.
The interaction (i.e., the connection) targets that neuron, which
(i) is in the direction of the arrow, or (ii) is at the circle, or
(iii) is targeted by a connection being targeted by the small
circle, and so on. The synapse could be of two basic types here,
it is either additive or multiplying type. An additive synapse can
be either excitatory or inhibitory. We assume that only the
excitatory synapses can be adapted. At the same time, we also
assume that feedforward inhibition, which is always present
(\cite{buzsaki84feedforward}), plays a role and learning occurs
relative to a negative background. In turn, feedforward inhibitory
synapses -- which are not shown in our figures -- and adapting
excitatory synapses may exert an \textit{effective} inhibition.
That is, matrices representing excitatory synapse sets may have
negative elements.}\label{f:types}
\end{figure}

Connectionist systems are special non-linear systems having graph
like structures. Nodes of the graph are called neurons, whereas
directed edges are the connections. Figure \ref{f:types} depicts a
neuron subject to \emph{local interactions}. An interaction is
called local if it is exerted by a directed connection. The
targeted neuron is the subject of the interaction. The end of the
directed connection is depicted by a small circle, called synapse.
This synapse could be of three types here, it is either
excitatory, inhibitory or of multiplying type. The first two of
these types are widely considered as neuronal operations.
Multiplying type synapse is, however, also possible: a neuron may
affect another neuron by modulating the `gain' of that neuron in a
multiplicative way (\cite{salinas00gain}). For a review on
different processing capabilities of single neurons, see, e.g.,
\cite{koch00therole}).

From the point of view of neuronal modelling,
\begin{enumerate}
    \item the series of observations constitute the \textit{input} to the model,
    \item the \textit{learning task} corresponds to finding the best parametrization
    and the best hidden variables given all past observations and subject to constraints
    prescribed by the norm (the measure) of a \textit{model} and the noise of assumed
    by this model, whereas
    \item \textit{filtering} corresponds to the estimation of the hidden variables
    given the past observations. \label{enum:filter}
\end{enumerate}

The seminal work of Hebb (\cite{Hebb49Organization}) was a
brilliant first attempt in its time to link neurophysiology with
higher order behavioral phenomena studied by psychology. The
central thesis postulated that changes in synaptic connection
strength is primarily based on the correlation between the pre-
and postsynaptic neural activities. Recent experiments
(\cite{markram97regulation,magee97synaptically,bell97synaptic},
for a review, see, e.g., \cite{Abbott00Synaptic}) revealed,
however, that exact timing and temporal dynamics of the neural
activities play a crucial role in forming the neuronal base of
plasticity. The novel concept to generally describe the modified
variants of the original form of Hebbian learning is called
spike-time dependent synaptic plasticity (STDP). The term `spike'
denotes the fast potential change which propagates over the
neurons and induces synaptic neurotransmitter release, which, in
turn, may result in synaptic modifications. STDP means that
\textit{strengthening} occurs (i) if the neuron fires and (ii) if
the excitatory synapse targeting this neuron delivered a spike
within a narrow time window around the time of firing. On the
other hand, \textit{weakening} occurs if the delivered spike at
the excitatory synapse is outside of this short time window.

\subsection{Kalman-filter recursion} \label{ss:KFR}

Let us consider the following linear dynamical system (LDS):
\begin{eqnarray}
  \y_t &=& \Hb \x_t+\n_t  \text{ observation process} \label{e:megf din}\\
  \x_{t+1} &=& \F \x_t+\m_t  \text{ dynamics of hidden variables} \label{e:hidden din}
\end{eqnarray}
where $\m_t \propto \N(0,\Pi)$, $\n_t \propto \N(0,\Sigma)$ are
independent noise processes. Here, notation $\N(\m,\Sigma)$ is a
shorthand to denote a stochastic variable of expectation value
$\m$ and covariance matrix $\Sigma$. Our task is the estimation of

hidden variables $\x(t) \in \mathbf{R}^n$ given the series of
observations $\y(\tau) \in \mathbf{R}^p$, $\tau \leq t$.

For estimations in squared (Euclidean) norm and Gaussian noise,
the optimal solution was derived by Rudolf Kalman
(\cite{Bagchi93Optimal,Elliott95Hidden}). The Kalman-filter
recursion is reproduced here: Let $E$ and $Cov$ denote the
expectation value and the covariance matrix operators,
respectively. Let us introduce the following notations:
$\hat{\x}(t|\tau)=E(\x_t|\y_1,\ldots \y_{\tau})$,
$\mathbf{N}_t=Cov(\x_t|\y_1,\ldots \y_t)$, and
$\M_t=Cov(\x_t|\y_1,\ldots \y_{t-1})$

\begin{lem}[Kalman-filter recursion]
Assume that $\hat{\x}(t-1|t-1)$,$\mathbf{N}_{t-1}$ has been
determined. Then
\begin{eqnarray}
\hat{\x}(t|t-1) &=& \F\hat{\x}(t-1|t-1) \label{e:priorhidden}\\
\M_t &=& \F \mathbf{N}_{t-1} \F^T +\Pi \\
\K_t^f &=& \M_t\Hb^T(\Hb\M_t\Hb^T+\Sigma)^{-1} \label{e:Kalman gain}\\
\hat{\x}(t|t) &=& \hat{\x}(t|t-1)+
\K_t^f\big(\y_t-\Hb\F\hat{\x}(t-1|t-1)\big)\label{e:posteriorhidden}\\
\mathbf{N}_t &=& (\mathbf{I}-\K_t^f\Hb)\M_t \label{e:noise}
\end{eqnarray}
where $\mathbf{I}$ denotes the identity transformation (i.e., the
identity matrix) and superscript $T$ denotes transposition.
\end{lem}

Combining equations \ref{e:priorhidden} and
\ref{e:posteriorhidden}, the following \textit{filter-equation}
emerges:
\begin{eqnarray}
\hat{\x}(t+1|t+1) &=
\underbrace{\F\hat{\x}(t|t)}_{\hat{\x}(t+1|t)}+
\K_{t+1}^f\big(\y_{t+1}-\Hb\F\hat{\x}(t|t)\big) \label{e:filter}
\end{eqnarray}
One may also introduce the so called \textit{prediction equation}:
\begin{eqnarray}
\hat{\x}(t+1|t) &= \F\hat{\x}(t|t-1)+
\K_{t}^{p}\big(\y_{t}-\Hb\hat{\x}(t|t-1)\big)=\F\hat{\x}(t|t)
\label{e:prediction}
\end{eqnarray}
where $\K^p_t=\F\K_{t}^f$, which can be used to make an estimation
\textit{before} the $(t+1)$th measurement.

In the literature, the two different quantities, i.e., $\K_t^f$
(Eq.~\ref{e:Kalman gain}) and $\F\K_{t}^f$
(Eq.~\ref{e:prediction}) are both called Kalman-gains.

\subsection{Review of the recursive prediction error method}
\label{ss:RPE}

Here, we shall consider the filtering task \textit{and} the
learning task. This latter will be restricted to the learning of
parameters required for proper adjustment of the Kalman-gain.

Let us make the following notes. Iterations of $\K_t^f$, $\K_t^p$,
$\M_t$, and $\mathbf{N}_t$ are independent from measurements on
$\y_1, \ldots \y_t$. However, these quantities depend from
quantities $\{\Hb, \F, \Sigma$ and $\Pi\}$. The dependencies, at
first sight, do not seem to admit a neuronal form. The problem is
in the computation of the Kalman-gain (Eq.~\ref{e:Kalman gain}):
this equation includes a matrix inversion, which does not admit a
connectionist (i.e., artificial neuronal) network form. Moreover,
it seems unlikely, that previous knowledge of covariance matrices
$\Sigma$ and $\Pi$ can be assumed for neuronal systems.

In the procedure that follows, we shall assume that the
Kalman-gain constitutes the unknown parameter of the system. The
Kalman-gain will be estimated on-line using the measured values of
$\y_t$ (i.e., the input of the neuronal architecture). Because an
on-line estimation makes use of (i) the estimated parameters, (ii)
the optimization of the activities given the actual observations,
and (iii) updates the estimated parameters given the optimized
activities and the previous estimations, on-line methods are best
suited to changing world. The price is that on-line estimation may
not be optimal for all past observation (this is not desirable in
a changing world, anyway). Instead, on-line estimation becomes
optimal asymptotically. It is well known that under rather mild
conditions, both Kalman-gains, i.e., $\K_t^p$ ,$\K_t^f$ converge
with exponential speed to the asymptotic $\K^p$ and $\K^f$ values,
respectively (i.e., $\lim_t \K_t=\K$). In turn, on-line estimation
is an attractive alternative when the goal is the estimation of
the parameter(s) of the Kalman-gain. We shall substitute equations
\ref{e:priorhidden}-\ref{e:noise} by the on-line methods. We shall
apply the recursive prediction error method (\cite{Ljung83Theory})
to \textit{derive} recursive estimation for the Kalman-gain.

Let $\K=\K(\theta)$ denote the parametrization of $\K$, where
$\K(\theta)$ may depend on $\theta$ arbitrarily. We shall use this
freedom to choose a particular form of the dependence. Now, our
task is to estimate $\theta$ and then to compute $\K(\theta)$
using the estimated value of $\theta$. For the sake of simplicity,
variable $\theta$ will be scalar in the derivation below. The same
derivation follows for vector and matrix variables. Also,
derivation concerns the estimation of matrix $\K^p_t$, but similar
derivation can be provided for matrix $\K^f_t$.

First, the prediction equation (Eq.~\ref{e:prediction}) shall be
investigated. Let $\hat{\x}(t,\theta)$ denote the predictive
estimation of the hidden variables belonging to $\theta$. Equation
\ref{e:prediction} can be written as
\begin{eqnarray}
\hat{\x}(t+1,\theta)&=&\F\hat{\x}(t,\theta)+\K^p(\theta)(\y(t)-
\Hb\hat{\x}(t,\theta)) \label{e:x}
\end{eqnarray}
The goal is to find the value of $\theta$ that minimizes the
squared norm of the \textit{reconstruction error} vector, which is
defined as $\epsilon(t,\theta)=\y(t)- \Hb\hat{\x}(t,\theta)$.
Vector $\Hb\hat{\x}(t,\theta)$, which is derived from the hidden
variables and should match the input in squared norm, will be
called the \textit{reconstructed input}. By definition, the
reconstruction error vector is a stochastic variable with zero
mean and $\Lambda$ covariance matrix ($\N(0,\Lambda)$). For the
purpose of on-line estimation, the $\Lambda$ covariance matrix
needs to be estimated from the data. The said goal, in
mathematical terms, has the following form:
\begin{eqnarray}
V(\theta)=E\left(
\epsilon(t,\theta)^T\Lambda^{-1}\epsilon(t,\theta)\right) \, \,
\rightarrow \, \,\min_{\theta} \label{e:objective}
\end{eqnarray}
Equation \ref{e:objective} can be seen as a maximum likelihood
problem. We shall apply the following procedure: (i) The
expectation value will be estimated by sample averaging, i.e.,
stochastic gradient approximation will be used. (ii) $\theta$ will
be minimized along the gradient. Let us make use of the notation:
$\w(t,\theta)=\frac{\partial}{\partial\theta}\hat{\x}(t,\theta)$
and let $\hat{\y}(t,\theta)=\Hb\hat{\x}(t,\theta)$ denote the
reconstructed input. Minimization of Eq.~\ref{e:objective} leads
to: $\hat{\theta}(t) = \hat{\theta}(t-1) + \gamma(t) \vb(t)^T
\Lambda^{-1}\epsilon(t) $, where $\gamma(t)>0$ is the learning
rate, shorthand $\vb(t,\theta)$ is given as
$\vb(t,\theta)=\frac{\partial}{\partial\theta}(\hat{\y}(t,\theta)-\y(t))$
and a recursive equation can be derived for $\w(t+1,\theta)$ from
Eq.~\ref{e:x} if it is differentiated according to $\theta$. The
following recursive estimation can be gained for the Kalman-gain
matrix  $\K$:
\begin{lem}[Local Kalman-filter recursion]
Assume that $\hat{\x}(t)$, $\hat{\w}(t)$, $\hat{\theta}(t)$,
$\hat{\Lambda}^{-1}(t)$ are given. Then
\begin{eqnarray}
\hat{\y}(t)&=&\Hb\hat{\x}(t) \label{e:K1}\\
\epsilon(t)&=&\y(t)-\hat{\y}(t) \\
\hat{\x}(t+1)&=&\F\hat{\x}(t)+\K(\hat{\theta}(t))\epsilon(t)\\
\hat{\vb}(t)&=&\Hb\hat{\w}(t) \label{e:K3}\\
\hat{\w}(t+1)&=& \F\hat{\w}(t)+\K'(\hat{\theta}(t))\epsilon(t)-
\K(\hat{\theta}(t))\hat{\vb}(t) \label{e:w_hat}\\
\hat{\theta}(t+1)&=& \hat{\theta}(t) +
\gamma(t) \hat{\vb}^T(t) \hat{\Lambda}^{-1}(t) \epsilon(t) \label{e:theta}\\
\hat{\Lambda}(t+1)&=&\hat{\Lambda}(t)+\gamma(t)
\left[\epsilon(t)\epsilon(t)^T-\hat{\Lambda}(t)\right]
\label{e:K70}
\end{eqnarray}
\end{lem}
\noindent where $\K'$ denotes $\frac{\partial K}{\partial\theta}$.
The auxiliary vector $\w$, which can be derived from the hidden
vector $\hat{\x}$ by differentiation according to $\theta$, will
play a crucial role in providing neuronal representation. For
later purposes, we rewrite Eq.~\ref{e:K70} into the following
stable form:
\begin{eqnarray}
\hat{\Lambda}^{-1}(t+1)&=&\hat{\Lambda}^{-1}(t)+\gamma(t)
\left[\hat{\Lambda}^{-1}(t) -
\left(\hat{\Lambda}^{-1}(t)\epsilon(t)\right)\left(\hat{\Lambda}^{-1}(t)\epsilon(t)\right)^T
\right]\label{e:K7}
\end{eqnarray}
In both recursions, i.e., in Eq.~\ref{e:K70} and in
Eq.~\ref{e:K7}, matrix $\hat{\Lambda}$ is an estimation of the
covariance matrix of the reconstruction error vector $\epsilon$.
In recursion Eq.~\ref{e:K7}, the inverse of the correlation matrix
is estimated directly. Update Eq.~\ref{e:K70} is a
\textit{signal-Hebbian} learning rule. Update Eq.~\ref{e:K7} is in
different form and can be seen as a neural update if spike-time
dependent synaptic plasticity is taken into account, as it will be
described later.

\section{Local Kalman-filter} \label{s:LKF}

Now, let us choose an element of matrix $\K$, say $\K_{ij}$. In
what follows, we shall make the particular assumption that
$\K_{ij}$ is an exponential function of $\theta$. This assumption
allows us to simplify the architecture considerably. The
simplification is warranted by the particular property of the
exponential dependence that $\K'_{ij}=\K_{ij}$, which will be
important when the local connectionist architecture will be
presented: We can use matrix $\K$ for both purposes, i.e., for
$\K$ itself and for $\K'$. In our formulation, the recursive
correction of the Kalman-gain assumes the multiplying
\textit{exponentiated gradient} form
(\cite{littlestone95online,kivinen97additive}):
\begin{eqnarray}
\K_{ij}(\theta(t+1)) &=& \exp\left(\gamma(t) \hat{\vb}(t)
\hat{\Lambda}^{-1}(t) \epsilon(t)\right) \, \K_{ij}(\theta(t))
\end{eqnarray}
Now, Eq.~\ref{e:w_hat} can be written as
\begin{eqnarray}
\hat{\w}(t+1)&=& \F\hat{\w}(t)+\K(\hat{\theta}(t))\left(
\epsilon(t)- \hat{\vb}(t) \right)\\
&=& \F\hat{\w}(t)+\K(\hat{\theta}(t))\left( \epsilon(t)-
(\Hb\hat{\w}(t)) \right)\\
&=& \K(\hat{\theta}(t))\epsilon(t) + \left( \F -
\K(\hat{\theta}(t))\Hb \right)\hat{\w}(t)
 \label{e:w_hat2}
\end{eqnarray}

\subsection{Architecture} \label{ss:archi}

\begin{figure}[ht!]
\centering\includegraphics[width=10cm]{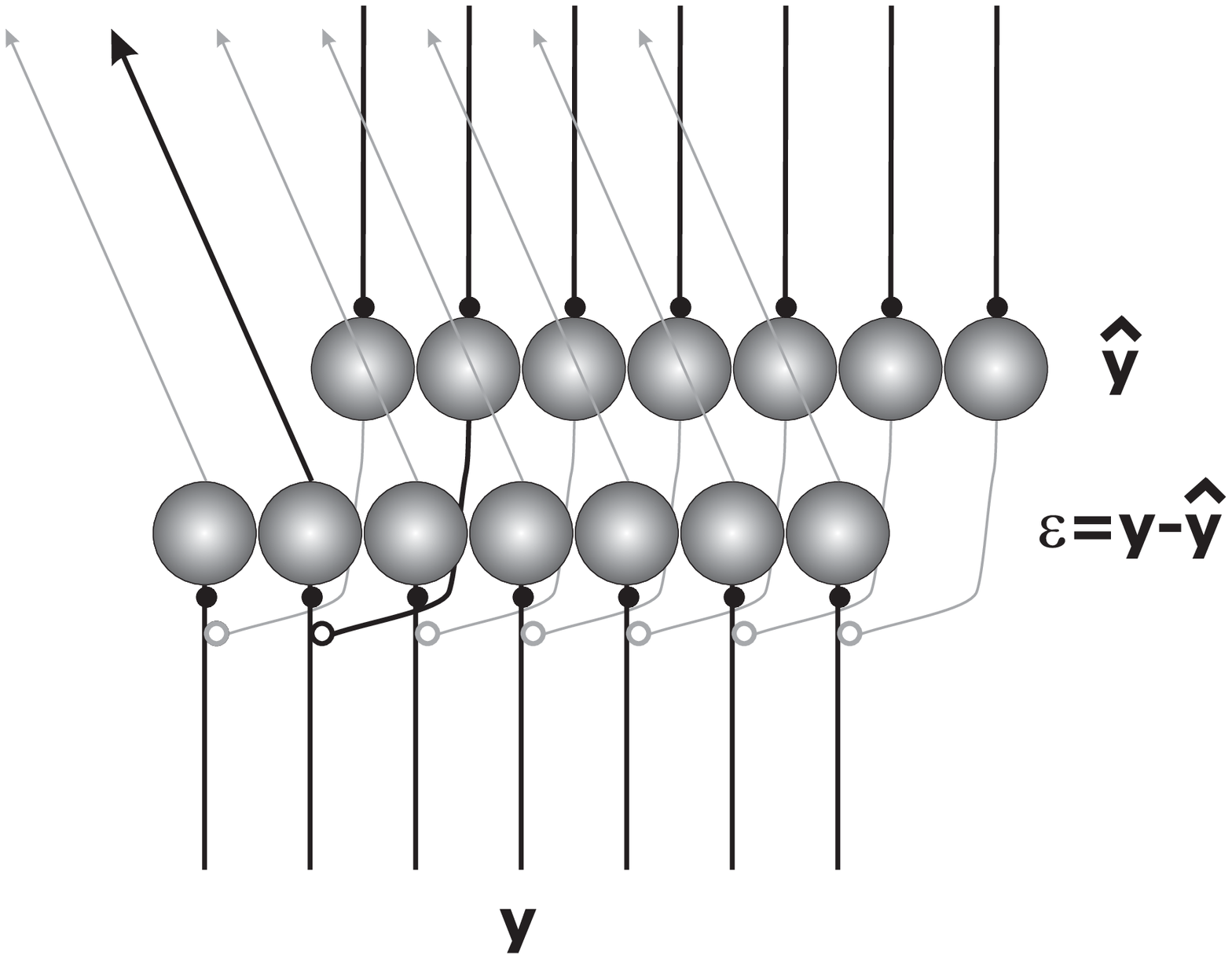}
\caption{\textbf{Computation of reconstruction error $\epsilon$}
\newline
Difference between two vector quantities can be computed by an
ordered array of connections of inhibitory type.}\label{f:diff}
\end{figure}
The reconstruction error vector ($\epsilon$) is computed by
differencing between the input ($\y$) and the estimated input
$\hat{\y}$. The circuitry is shown in Fig.~\ref{f:diff}.

\begin{figure}[h!]
\centering\includegraphics[width=10cm]{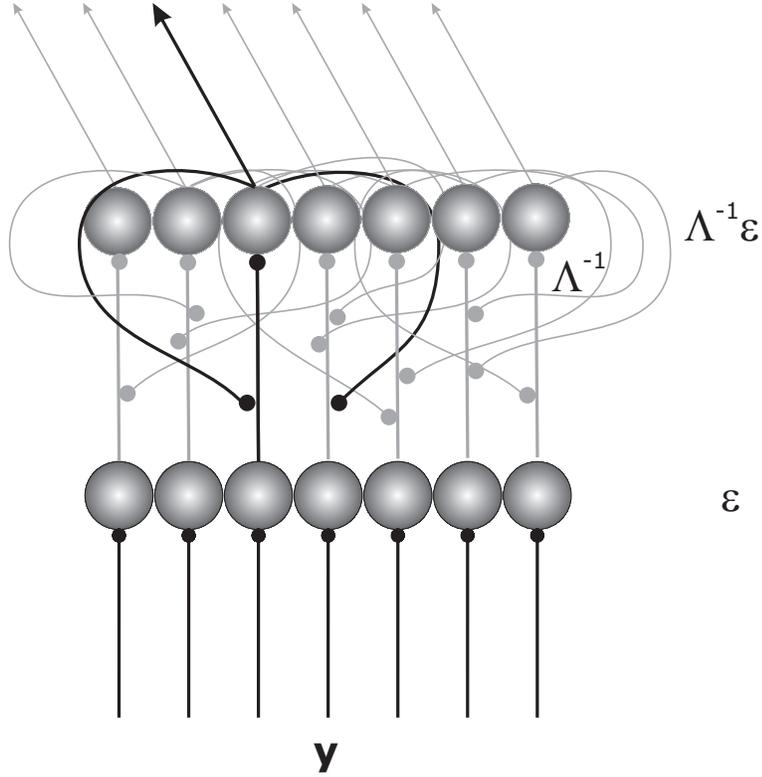}
\caption{\textbf{Learning and multiplying by the inverse of
correlation matrix $\Lambda$}
\newline
There is a \textit{recurrent collateral} system (a full
connectivity excitatory feedback structure) at the output layer.
(Only few connections are shown.) Noise is present and accidental
coincidences between excitatory inputs and neuronal firing can
give rise to self-strengthening proportional to the connection
strength \cite{lorincz00parahippocampal}. This effect provides the
first (positive) term of Eq.~\ref{e:K7}. We assume that the
recurrent collaterals have considerable delays and taht synapses
are weakened because of the spike-time dependent synaptic
plasticity process. The weakening effect is proportional to the
postsynaptic activities and to the activities carried by the
recurrent collaterals: This effect provides the negative term of
Eq.~\ref{e:K7}. In turn, the architecture executes the update of
Eq.~\ref{e:K7} and multiplies the input (i.e., $\epsilon$) by
$\Lambda^{-1}$. For details, see text.}\label{f:lambda_epsi}
\end{figure}

Computation of the parameter $\theta$ of the Kalman-gain can be
performed by computing vector $\Lambda^{-1}\epsilon$. To this end,
vector $\epsilon$ is to be multiplied by the inverse of the
correlation matrix $\Lambda$ of the $\epsilon$ vectors themselves.
Matrix $\Lambda$ is to be learned by the RPE method.

Learning can be accomplished by Hebbian means as depicted in
Fig.~\ref{f:lambda_epsi}:  The reconstruction error layer provides
input for an auxiliary neural sub-layer. This layer will output
vector $\Lambda^{-1}\epsilon$. There is a full connectivity
excitatory feedback structure at this sub-layer. It is assumed
that noise is present in the system and that this noise gives rise
to self-strengthening for all feedback connections. This
self-strengthening effect is assumed to be proportional to the
connection strength, i.e., it is responsible for the first
(positive) term of Eq.~\ref{e:K7}. It is assumed, too, that this
feedback structure, called \textit{recurrent collaterals} have
considerable delays, feedback arrives to the excitatory synapses
late and weakening of the synapses occur because of the STDP
process. The weakening is proportional to the postsynaptic
activity and to the activity carried by the recurrent collateral
and, in turn, the negative term of Eq.~\ref{e:K7} emerges. The
emerging full update corresponds to Eq.~\ref{e:K7}.

For proper outputs, however, an additional structure is required:
The excitatory feedback effect should act only once. We assume
that an augmenting set of inhibitory neurons exist (these are not
shown in Fig.~\ref{f:lambda_epsi}) and that these inhibitory
neurons are excited by the same auxiliary neural sub-layer.
Moreover, it is assumed that connections originated by the
inhibitory neurons target the feedback excitatory synapses of the
auxiliary neural sub-layer. The extra step to excite an inhibitory
layer makes the inhibitory effect delayed relative to the
excitatory effect at the feedback excitatory synapses. Inhibition
stops the excitatory effect and excitatory feedback can occur only
once in each iteration step. It is intriguing that this complex
excitatory-inhibitory structure does exist in the cortex
(\cite{freund96interneurons}). It then follows that (i) the
circuitry of Fig.~\ref{f:lambda_epsi} performs the RPE computation
prescribed by Eq.~\ref{e:K7} and that the output of the sub-layer
is equal to $\Lambda^{-1}\epsilon$.

\begin{figure}[h!]
\centering\includegraphics[width=10cm]{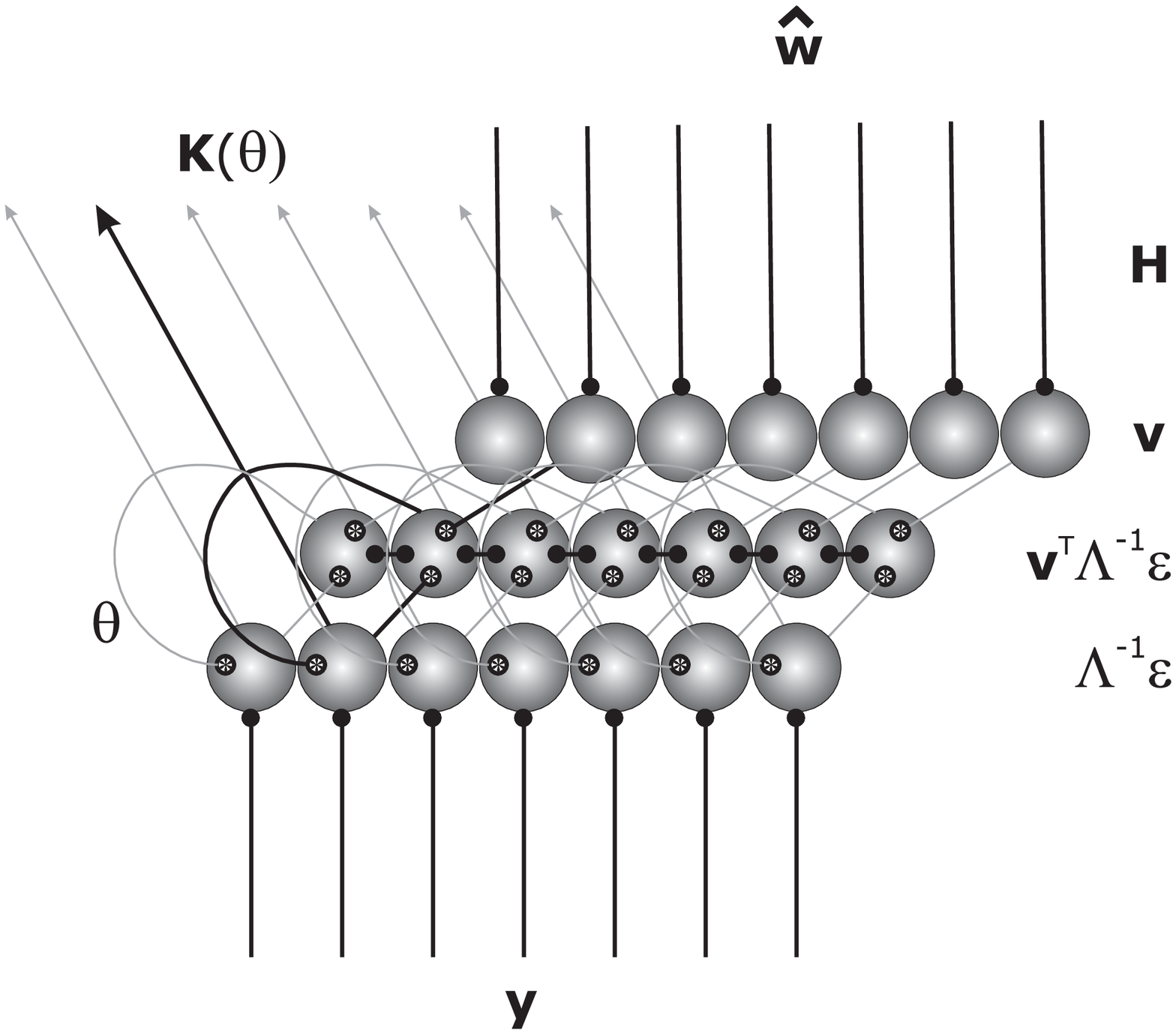}
\caption{\textbf{Computation and influence of parameter $\theta$,
the parameter of the Kalman-gain}
\newline
Component-by-component multiplication of vector $\vb$
(Eq.~\ref{e:K3}) and vector $\Lambda^{-1}\epsilon$ are performed
and the components are summed up in a separate sub-layer. The
output of this layer influences the Kalman-gain via synapses of
multiplying types. These synapses influence the outputs of neurons
of the reconstruction error vector.}\label{f:kalman_gain}
\end{figure}

Last, the parameter of the Kalman-gain is to be computed. The
computation is made of three parts. First, vector $\vb$
(Eq.~\ref{e:K3}) and vector $\Lambda^{-1}\epsilon$ are to be
multiplied  component-by component. Then the individual terms need
to sum up to correct the previous estimation of the parameter of
the Kalman-gain. Finally, the re-estimated parameter influences
each outgoing connections in a uniform but non-linear fashion.
This non-linear dynamic effect is the attenuation process of the
Kalman-filter, i.e., the tuning of the Kalman-gain. We made the
assumption that this non-linear influence is, in fact,
exponential. The corresponding circuitry is shown in
Fig.~\ref{f:kalman_gain}. Although, here a connectionist model is
constructed, still, it may be important to emphasize that the
Kalman-gain influences the output of the reconstruction error
layer in a \textit{uniform} fashion. It is then possible that not
a full layer, but \textit{a single} (or a few) neuron(s) (i)
compute the scalar product of vectors $\vb$ and
$\Lambda^{-1}\epsilon$ and (ii) target the neurons of the
reconstruction error layer to tune the Kalman-gain as required for
Kalman-filtering.

\begin{figure}[h!]
\centering\includegraphics[width=8cm]{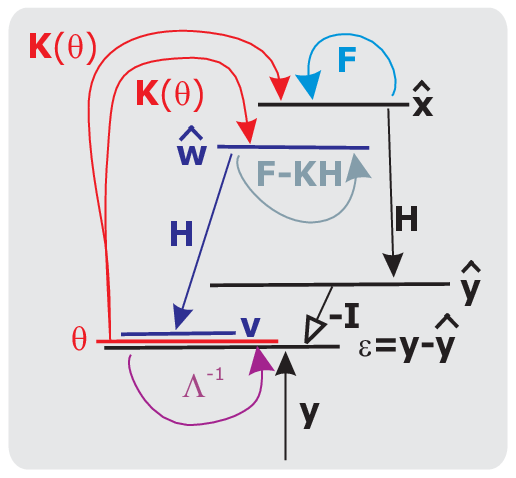}
\caption{\textbf{Architecture to execute the recursive prediction
error method}
\newline
The exponentiated dependence on the parameter of the Kalman-gain
allows for a simple loop structure. The computations of
Figs.~\ref{f:diff}-\ref{f:kalman_gain} together, perform
Kalman-filtering in the depicted loop as prescribed by
Eqs.~\ref{e:K1}-\ref{e:w_hat2}. Solid arrows: excitatory
connections. Empty arrow: inhibitory connections. Recursive
prediction error method proceeds as follows: Neural activities are
given at time $t$ at each layer. To compute the estimation at time
$t+1$, all activities are propagated through the connections
represented by the arrows of the figure, simultaneously. Emerging
constraints are as follows: (i) There should be an associative
matrix connecting components of the estimation of the hidden
variables and this matrix should be equal to $\F$. (ii) There
should also be an associative matrix connecting components of the
estimation of auxiliary vector $\w$ and this matrix should be
equal to $\F-\K\Hb$.}\label{f:KF_archi}
\end{figure}

The distinct parts of the RPE computation can be made recursive by
means of a loop architecture. The loop architecture is depicted in
Fig.~\ref{f:KF_archi}.

Figures \ref{f:diff}-\ref{f:KF_archi} demonstrate that
Kalman-filtering can be performed by local means in a loop through
the RPE method. This observation supports top-down modelling
efforts, which have appeared recently in the literature
(\cite{rao97dynamic,rao99predictive,bousquet99is,lorincz00parahippocampal,lorincz02mystery,todorov02optimal}).
The top-down modelling suggestion -- that Kalman-filtering may
play a role in cortical computations -- calls for reinforcement
from the bottom-up modelling methods of computational
neuroscience. Our work makes a step towards this direction; we
have transformed the suggestions of top-down modelling into a
dynamical neural architecture with appealing Hebbian learning
rules. The particular form of Kalman-filtering we presented here
poses several questions. We list a few of them. (i) How can be
this architecture mapped onto the neocortex? (ii) How can be the
derived computations performed in the neural substrate? (iii) In
particular, how reasonable is our assumption that the Kalman-gain
exponentially depends on its parameter? These questions, which can
be seen as the predictions of our model, remain open and call for
further studies.

\section{Conclusions} \label{s:conc}

In this paper we have presented a neural (connectionist)
representation for Kalman-filtering. The issue is worth to
consider given the recent advances in the literature
(\cite{rao99predictive,lorincz02mystery,todorov02optimal}).
Regarding the cited works we feel that all have the problem in
common: Kalman-filtering should be given a local connectionist
architecture for the proposed top-down approaches. Here, we have
shown that yes, connectionist representation is possible for
Kalman-filtering: The recursive prediction error method suits the
requirements imposed by Hebbian constraints. Moreover, RPE
provides an appealing on-line learning scheme. This is a promising
start. However, the mapping of the architecture to neocortical
areas and the question if the predictive matrices of the
Kalman-filter could be learned by Hebbian means remained open.
Also, further studies starting from parameters of the neuronal
substrate are in need to explore the particular predictions of the
RPE model of Kalman-filtering.

\section*{Acknowledgements}

Enlightening discussions with Gy\"orgy Buzs\'aki are gratefully
acknowledged. We would like to thank to G\'abor Szirtes his
valuable comments regarding the manuscript. This work was
supported by the Hungarian National Science Foundation (Grant No.
OTKA 32487).

\bibliography{Local_KF_gain_arxiv}

\begin{thebibliography}{10}
\expandafter\ifx\csname url\endcsname\relax
  \def\url#1{\texttt{#1}}\fi
\expandafter\ifx\csname urlprefix\endcsname\relax\def\urlprefix{URL }\fi

\bibitem{rao97dynamic}
R.~Rao, D.~Ballard, Dynamic model of visual recognition predicts neural
  response properties in the visual cortex, Neural Comput 9 (1997) 721--763.

\bibitem{rao99predictive}
R.~Rao, D.~Ballard, Predictive coding in the visual cortex: \mbox{A} functional
  interpretation of some extra-classical receptive-field effects, Nature
  Neuroscience 2 (1999) 79--87.

\bibitem{bousquet99is}
O.~Bousquet, K.~Balakrishnan, V.~Honavar, Proceedings of the Pacific Symposium
  on Biocomputing, 1999, Ch. Is the Hippocampus a Kalman Filter?, pp. 619--630.

\bibitem{todorov02optimal}
E.~Todorov, M.~Jordan, Optimal feedback control as a theory of motor
  coordination, Nature Neuroscience 5 (2002) 1226--1235.

\bibitem{lorincz00parahippocampal}
A.~L{\H o}rincz, G.~Buzs\'aki, The parahippocampal region: \mbox{I}mplications
  for neurological and psychiatric dieseases, in: H.~Scharfman, M.~Witter,
  R.~Schwarz (Eds.), Annals of the New York Academy of Sciences, Vol. 911, New
  York Academy of Sciences, New York, 2000, Ch. Two--phase computational model
  training long--term memories in the entorhinal--hippocampal region, pp.
  83--111.

\bibitem{lorincz02mystery}
A.~L{\H o}rincz, B.~Szatm\'ary, G.~Szirtes, Mystery of structure and function
  of sensory processing areas of the neocortex: \mbox{A} resolution, J. Comp.
  Neurosci. 13 (2002) 187–205.

\bibitem{lorincz02relating}
A.~L{\H o}rincz, G.~Szirtes, B.~Tak\'acs, I.~Biederman, R.~Vogels, Relating
  priming and repetition suppression, Int. J. of Neural Systems 12 (2002)
  187--202.

\bibitem{keri02categories}
S.~K\'eri, G.~Benedek, Z.~Janka, P.~Aszal\'os, B.~Szatm\'ary, G.~Szirtes,
  A.~L{\H o}rincz, Categories, prototypes and memory systems in alzheimer's
  disease, Trends in Cognitive Science 6 (2002) 132--136.

\bibitem{Ljung83Theory}
L.~Ljung, T.~Soderstrom, Theory and practice of recursive identification, MIT
  Press, Cambridge, MA, 1993.

\bibitem{salinas00gain}
E.~Salinas, P.~Thier, Gain modulation: a major computational principle of the
  central nervous system., Neuron 27 (2000) 15--21.

\bibitem{koch00therole}
C.~Koch, I.~Segev, The role of single neurons in information processing, Nature
  Neuroscience 3 (2000) 1171--1177.

\bibitem{buzsaki84feedforward}
G.~Buzs\'aki, Feed-forward inhibition in the hippocampal formation, Prog.
  Neurobiol. 22 (1984) 131--153.

\bibitem{Hebb49Organization}
D.~Hebb, The Organization of Behavior: A Neuropsychological Theory, Wiley, New
  York, 1949.

\bibitem{markram97regulation}
H.~Markram, J.~Lubke, M.~Frotscher, B.~Sakmann, Regulation of synaptic efficacy
  by coincidence of postsynaptic \mbox{AP}s and \mbox{EPSP}s, Science 215
  (1997) 213--215.

\bibitem{magee97synaptically}
J.~Magee, D.~Johnston, A synaptically controlled, associative signal for
  hebbian plasticity in hippocampal neurons, Science 275 (1997) 209--213.

\bibitem{bell97synaptic}
C.~Bell, V.~Han, Y.~Sugawara, K.~Grant, Synaptic plasticity in a
  cerebellum-like structure depends on temporal order, Nature 387 (1997)
  278--281.

\bibitem{Abbott00Synaptic}
L.~Abbott, S.~Nelson, Synaptic plasticity: taming the beast, Nature
  Neuroscience 3 (2000) 1178--1183.

\bibitem{Bagchi93Optimal}
A.~Bagchi, Optimal control of stochastic systems, Prentice Hall, New York,
  1993.

\bibitem{Elliott95Hidden}
R.~Elliott, A.~Lakhdar, J.~Moore, Hidden \mbox{M}arkov models.
  \mbox{E}stimation and control, Springer Verlag, New York, 1995.

\bibitem{littlestone95online}
N.~Littlestone, P.~Long, M.~Warmuth, On-line learning of linear functions,
  Journal of Computational Complexity 5 (1995) 1--23.

\bibitem{kivinen97additive}
J.~Kivinen, M.~K. Warmuth, Additive versus exponentiated gradient updates for
  linear prediction, Information and Computation 5: 1-23, 132 (1997) 1--64.

\bibitem{freund96interneurons}
T.~Freund, G.~Buzs\'aki, Interneurons of the hippocampus, Hippocampus 6 (1996)
  345--470.

\end{thebibliography}
\bibliographystyle{elsart-num}
\end{document}